\documentclass{article} 
\usepackage{iclr2027_conference,times}

\usepackage{amsmath,amsfonts,bm}

\def\eqref#1{equation~\ref{#1}}

\def\1{\bm{1}}

\def\vmu{{\bm{\mu}}}

\def\vc{{\bm{c}}}

\def\vp{{\bm{p}}}

\def\vr{{\bm{r}}}

\def\vx{{\bm{x}}}

\def\vz{{\bm{z}}}

\def\mSigma{{\bm{\Sigma}}}

\DeclareMathAlphabet{\mathsfit}{\encodingdefault}{\sfdefault}{m}{sl}
\SetMathAlphabet{\mathsfit}{bold}{\encodingdefault}{\sfdefault}{bx}{n}

\newcommand{\R}{\mathbb{R}}

\usepackage{hyperref}
\usepackage{url}

\usepackage{booktabs}
\usepackage{multirow}
\usepackage{graphicx}
\usepackage{xspace}
\usepackage{wrapfig}
\usepackage{enumitem}

\newcommand{\method}{Res-SCBMs\xspace}
\newcommand{\singularmethod}{Res-SCBM\xspace}

\title{Residual Stochastic Concept Bottleneck Models}

\author{Stephen Marcou\thanks{Correspondence to \texttt{smarcou@ethz.ch}} \\
ETH Zurich
\And
Mateo Espinosa Zarlenga \\
University of Oxford
\And
Sonia Laguna \\
ETH Zurich
\AND
Moritz Vandenhirtz \\
ETH Zurich
\And
Julia E. Vogt \\
ETH Zurich
\And
Mateja Jamnik \\
University of Cambridge
}

\iclrfinalcopy 
\begin{document}

\maketitle

\begin{abstract}
Concept-based models (CMs) learn human-interpretable intermediate representations that enable users to understand the model's reasoning and steer its predictions through test-time concept interventions. However, in realistic settings, their performance can be limited by concept incompleteness, where the training concepts fail to capture all task-relevant information. Adding an unconstrained residual pathway to the model's predictive process can overcome the consequences of incompleteness. However, this introduces an inscrutable, leak-prone prediction pathway that may weaken concept interventions, as these often leave residual variables unchanged. To address this limitation, we introduce Residual Stochastic Concept Bottleneck Models (\method), a family of concept-based models that jointly model concept and residual variables through a multivariate Gaussian distribution. \method learn a covariance structure that explicitly captures dependencies between the known concepts and learnt residual variables, allowing interventions to propagate to statistically dependent residual variables through Gaussian conditioning. This extends the intervention capabilities of previous CMs 
without requiring concept completeness during training.
Our results show that, across multiple concept-incomplete benchmarks, \method improve predictive performance over competing baselines while retaining effective test-time concept interventions.\looseness-1

\end{abstract}

\section{Introduction}
As deep learning models are increasingly deployed across a wide range of domains, understanding the reasoning underlying their predictions has become more important. Concept-based models (CMs) have emerged as a promising approach to incorporate interpretability by design \citep{DBLP:conf/icml/KohNTMPKL20, DBLP:conf/nips/ZarlengaBCMGDSP22, DBLP:conf/iclr/YuksekgonulW023, DBLP:conf/nips/VandenhirtzLMV24}. Rather than predicting the task output $\hat{y}$ directly from the input $\vx$, CMs first predict a set of human-interpretable concept representations $\hat{\vc} = g(\vx)$, which are then used to produce the final prediction $\hat{y} = f(\hat{\vc})$. These intermediate representations let users inspect the concepts a model uses in its reasoning and, importantly, intervene on intermediate concepts at test time. Through these interventions, users can correct mispredicted concepts in the model's intermediate reasoning and communicate high-level hints as concepts \citep{DBLP:conf/nips/LagunaMVV24, DBLP:conf/aaai/ChauhanTFSD23, DBLP:conf/nips/ZarlengaCDWSJ23}.

A central challenge for CMs is \emph{concept incompleteness} \citep{DBLP:conf/nips/YehKALPR20, DBLP:journals/corr/abs-2106-13314}, 
where the predefined concept set fails to capture all information relevant to the downstream task. In practice, constructing a complete concept set 
can be difficult, and obtaining reliable concept annotations may require substantial effort, whether through manual labelling, domain expertise, or automated annotation pipelines \citep{DBLP:conf/iclr/OikarinenDNW23, DBLP:conf/iclr/YuksekgonulW023}. As a result, models constrained to operate solely through the concept bottleneck may suffer from limited predictive performance \citep{DBLP:conf/nips/ZarlengaBCMGDSP22,DBLP:conf/nips/HavasiPD22, DBLP:conf/iclr/YuksekgonulW023}.

To address this limitation, several methods \citep{DBLP:conf/iclr/YuksekgonulW023, DBLP:conf/cvpr/ShangZZNYW24} have explored augmenting an interpretable concept representation with additional residual representations designed to capture task-relevant information not explained by the predefined concepts. However, introducing a residual channel creates a tension between predictive capacity and intervenability \citep{DBLP:conf/nips/ZarlengaBCMGDSP22, DBLP:journals/corr/abs-2504-14094}. To address concept incompleteness, the residual representation must carry task-relevant information. Yet, because this provides an alternative pathway to the downstream predictor, concept corrections may have less influence on the final prediction \citep{DBLP:journals/corr/abs-2606-10669, DBLP:journals/corr/abs-2504-14094}. Existing work has shown that additional latent representations can coexist with effective concept interventions \citep{DBLP:conf/nips/HavasiPD22, DBLP:journals/corr/abs-2606-10669}, but these representations are typically treated as separate information pathways. In particular, intervening on a predefined concept does not explicitly update the residual representation to reflect dependencies between the two, potentially leaving correlated residual information unchanged after the intervention. A key challenge is therefore not only to retain intervenability with a residual channel, but to ensure that the residual representation itself responds consistently when concepts are corrected.

\begin{figure}[!t]
    \centering
    \includegraphics[width=\linewidth]{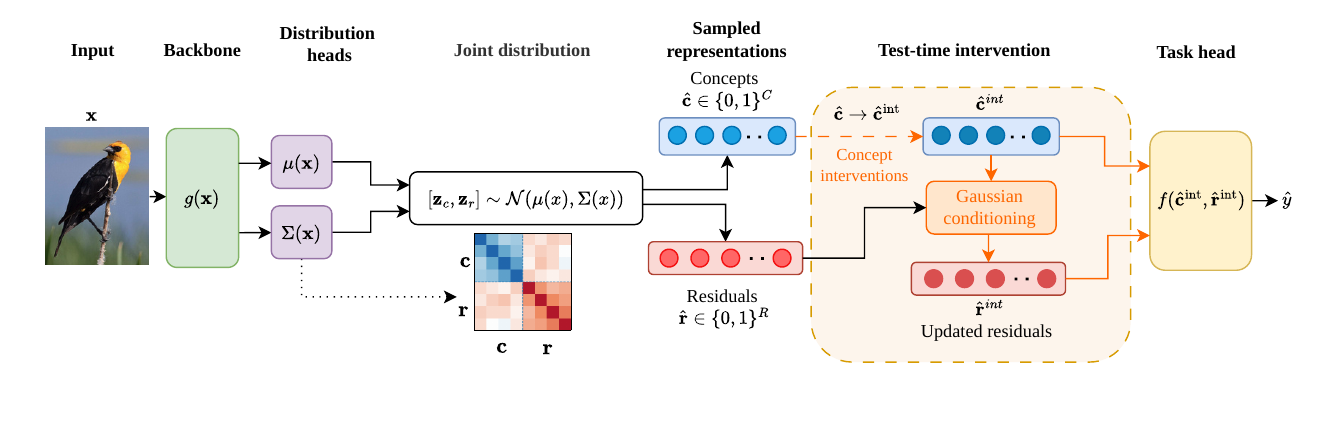}
    \vspace*{-10mm}
    \caption{\textbf{Overview of \method.} Concept and residual variables are jointly modelled through a multivariate Gaussian distribution. At test time, concept interventions propagate to statistically dependent residual variables through Gaussian conditioning, updating the residual representation before prediction.\looseness-1}
    \label{fig:res_scbm}
    \vspace{-1.5mm}
\end{figure}

To address this, we introduce \emph{Residual Stochastic Concept Bottleneck Models} (\method), a concept-based architecture that introduces a residual channel designed to capture task-relevant information missing from the predefined concept set while preserving effective concept interventions (Figure~\ref{fig:res_scbm}). Rather than modelling concepts and residuals independently, \method draw inspiration from Stochastic Concept Bottleneck Models (SCBMs) \citep{DBLP:conf/nips/VandenhirtzLMV24} and represent concepts and residuals jointly via a multivariate Gaussian distribution, whose covariance structure captures dependencies between the two representations. Consequently, concept interventions, both on known concepts and on the residual variables, can propagate through Gaussian conditioning to other statistically dependent variables, which may include residuals. We achieve this by introducing an intervention-aware training objective, inspired by \citet{DBLP:journals/corr/abs-2606-10669}, that encourages the model to retain predictive performance after concept correction. Our results suggest that
\method recover the predictive performance lost by competing approaches such as SCBMs under concept incompleteness, while retaining effective concept interventions.

Therefore, our main contributions can be summarised as follows:
\begin{itemize}[leftmargin=9pt,topsep=0pt,itemsep=-1pt]
\item \method, a probabilistic concept-residual architecture that retains task-relevant information under concept incompleteness and propagates concept interventions to statistically dependent residual variables.
\item An empirical evaluation demonstrating that \method improve predictive performance on concept-incomplete tasks while maintaining effective concept interventions across different residual dimensionalities.
\end{itemize}

\section{Background \& Related Work}

\paragraph{Concept Bottleneck Models.} Most CMs can be expressed as a Concept Bottleneck Model (CBM) \citep{DBLP:conf/icml/KohNTMPKL20}. A CBM consists of two functions, a concept predictor $g(\mathbf{x})$ and a task predictor $f(\hat{\mathbf{c}})$, which are typically parameterised by neural networks. Given an input $\vx \in \R^D$, the concept predictor maps $\vx$ to a vector of $C$ human-interpretable concepts, $\hat{\vc} = g(\vx) \in [0,1]^C$. The task predictor then takes the predicted concept representation as input and produces the final task prediction, $\hat{y} = f(\hat{\vc})$. CBMs are trained on datasets of the form $\mathcal{D} = \{(\vx_i,\vc_i,y_i)\}_{i=1}^{N}$, where $\vx_i \in \R^D$ are the raw input features, $\vc_i \in \{0,1\}^C$ is the corresponding vector of human-interpretable concept annotations, and $y_i \in \mathcal{Y}$ is the downstream task label. While these concept annotations are traditionally provided by human annotators, recent variants obtain concepts automatically through label-free or unsupervised concept discovery pipelines \citep{DBLP:conf/iclr/OikarinenDNW23,DBLP:journals/corr/abs-2407-03921}. Several extensions of the standard CBM formulation have been proposed, including post-hoc CBMs, which construct concept representations from pretrained models  \citep{DBLP:conf/iclr/YuksekgonulW023}, probabilistic CBMs \citep{DBLP:conf/nips/VandenhirtzLMV24}, models with richer concept representations \citep{DBLP:conf/nips/ZarlengaBCMGDSP22}, and hybrid or residual CBMs that introduce additional capacity beyond the predefined concept bottleneck \citep{DBLP:journals/corr/abs-2106-13314,DBLP:conf/cvpr/ShangZZNYW24}.

\paragraph{Modelling concept dependencies.} A limitation of conventional CBMs is that they do not explicitly model dependencies between concepts \citep{DBLP:conf/nips/VandenhirtzLMV24, DBLP:conf/iclr/DominiciBZTGML25}. Such dependencies naturally arise in real-world concept sets. For example, \texttt{CUB} \citep{WahCUB_200_2011}, a popular benchmark for CMs, contains attributes describing related regions of a bird, including breast colour and underparts colour, which may exhibit strong statistical dependencies. Ignoring these dependencies can make test-time interventions less efficient \citep{DBLP:conf/eccv/SinghiKRA24}, as correcting one concept does not automatically update other related concepts, potentially requiring users to correct them individually. Several approaches account for such dependencies. ECBMs \citep{DBLP:conf/iclr/XuQMWL24} model dependencies between inputs, concepts, and task labels through a joint energy-based formulation, allowing concept corrections to affect related concepts. Causal approaches instead explicitly model directed relationships between concepts. For example, Causal Concept Graph Models \citep{DBLP:conf/iclr/DominiciBZTGML25} organise concepts according to a causal graph, allowing interventions to propagate through the modelled causal structure. Stochastic CBMs (SCBMs) \citep{DBLP:conf/nips/VandenhirtzLMV24} instead model the concepts jointly using a multivariate Gaussian distribution. Its covariance structure captures dependencies between concepts, allowing interventions on one concept to propagate to correlated concepts through Gaussian conditioning.\looseness-1

\paragraph{Residual capacity in CBMs.} To address concept incompleteness, several approaches have explored increasing the information capacity of concept-based models beyond the predefined concept set. Hybrid CBMs \citep{DBLP:journals/corr/abs-2106-13314} augment the interpretable concept bottleneck with an additional residual representation, allowing task-relevant information not captured by the concepts to bypass the bottleneck. Similarly, Hybrid Post-hoc Concept Bottleneck Models (PCBM-h) \citep{DBLP:conf/iclr/YuksekgonulW023} introduce a residual predictor that operates on the original backbone representation and adds its output to the concept-based prediction, allowing information missing from the concept bank to contribute to the downstream task. Incremental Residual Concept Bottleneck Models \citep{DBLP:conf/cvpr/ShangZZNYW24} introduce residual directions to capture missing information and subsequently attempt to associate them with additional human-interpretable concepts. In contrast, Concept Embedding Models (CEMs) \citep{DBLP:conf/nips/ZarlengaBCMGDSP22} increase the capacity of the concept bottleneck itself by representing each concept with a high-dimensional embedding rather than a single scalar value. This richer representation retains task-relevant information beyond binary concept labels while preserving a concept-structured bottleneck. Increasing information capacity beyond the predefined concepts can increase the risk of leakage, potentially reducing both interpretability and intervenability \citep{DBLP:journals/corr/abs-2504-14094}. Rather than treating all leakage as undesirable, \citet{DBLP:journals/corr/abs-2606-10669} distinguishes between harmful and benign leakage, showing that benign leakage can retain additional task-relevant information while preserving effective concept interventions.

Despite these advances, existing work has largely considered these two directions separately, either modelling dependencies within the concept representation or introducing additional capacity for information missing from the concept set. Existing methods do not jointly model concepts and residual variables in a way that allows concept interventions to propagate to statistically dependent residual variables. Our work addresses this limitation by jointly modelling concepts and residual variables, allowing interventions to propagate between them, while an intervention-aware training objective encourages the model to remain responsive to concept corrections.

\section{Residual Stochastic CBMs}
\label{sec:model-formulation}

We introduce Residual Stochastic Concept Bottleneck Models (Res-SCBMs),
a class of CMs that extend SCBMs with residual representations to capture task-relevant information not encoded by the observed concepts. Concepts and residuals are modelled jointly through a multivariate Gaussian, whose covariance structure captures dependencies between the two representations (see Figure~\ref{fig:res_scbm}).


\subsection{Model Formulation}
Given an input $\vx \in \mathbb{R}^D$, a \singularmethod models predicted concept logits $\vz_c \in \mathbb{R}^C$ and residual logits $\vz_r \in \mathbb{R}^R$ through a joint multivariate Gaussian distribution parameterised as:
\begin{equation}
    \label{eq:joint_distribution}
    \begin{bmatrix}
    \vz_c \\
    \vz_r
    \end{bmatrix}
    \sim
    \mathcal{N}\!\left(
    \begin{bmatrix}
    \vmu_c(\vx) \\
    \vmu_r(\vx)
    \end{bmatrix},
    \begin{bmatrix}
    \mSigma_{cc}(\vx) &
    \mSigma_{cr}(\vx) \\
    \mSigma_{cr}(\vx)^\top &
    \mSigma_{rr}(\vx)
    \end{bmatrix}
    \right).
\end{equation}

Here, the cross-covariance $\mSigma_{cr}$ captures statistical dependencies between the concept and residual representations. For \method to be most effective, residual variables should be neither independent of the concepts nor merely redundant with them. If residual variables are independent of concepts, concept interventions induce little or no change in the residual representation through Gaussian conditioning. Conversely, if residual variables are too strongly dependent on the concepts, they may largely replicate information already captured by the concept representation. Ideally, the residual variables capture complementary task-relevant information missing from the observed concept set while retaining  dependencies with the concepts that allow interventions to propagate to the residual representation.

Following \citet{DBLP:conf/nips/VandenhirtzLMV24}, the sampled
logits are mapped to stochastic binary concept and residual representations by
applying a sigmoid followed by Bernoulli sampling:
\begin{equation}
\begin{bmatrix}
\hat{\vc} \\
\hat{\vr}
\end{bmatrix}
\sim
\operatorname{Bernoulli}\!\left(
\sigma\!\left(
\begin{bmatrix}
\vz_c \\
\vz_r
\end{bmatrix}
\right)
\right)
\in \{0,1\}^{C+R}.
\label{eq:binary_representation}
\end{equation}
In practice, we estimate gradients during optimisation using a straight-through estimator. The concept and
residual representations are concatenated and passed jointly through the same
task head, $\hat{y}=f\!\left([\hat{\vc},\hat{\vr}]\right)$, where $[\cdot,\cdot]$ denotes vector concatenation.

\subsection{Concept interventions}
Modelling concepts and residuals jointly is central to \method.
The learned covariance structure allows an intervention on a concept to update not only correlated concepts, as in SCBMs \citep{DBLP:conf/nips/VandenhirtzLMV24}, but also residual
variables that are statistically dependent on the intervened concept. Following the intervention procedure of SCBMs and logit-based CBMs~\citep{DBLP:conf/icml/KohNTMPKL20}, we map binary concept corrections using empirical percentiles of the predicted concept probabilities on the training set. A corrected concept with $c_i^{\mathrm{int}}=0$ is mapped to the 5th percentile, whereas $c_i^{\mathrm{int}}=1$ is mapped to the 95th percentile. The corresponding intervention logit is given by
\begin{equation}
p_{i}^{\mathrm{int}}
=
(1-c_i^{\mathrm{int}})q_{0.05}^{(i)}
+
c_i^{\mathrm{int}}q_{0.95}^{(i)},
\qquad
z_{c,i}^{\mathrm{int}}
=
\sigma^{-1}\!\left(p_i^{\mathrm{int}}\right),
\label{eq:intervention_logits}
\end{equation}
where $q_{0.05}^{(i)}$ and $q_{0.95}^{(i)}$ denote the 5th and 95th percentiles of the empirical distribution of predicted probabilities for concept $i$ over the training set.

Notice then that when all concepts are intervened on with corrected concept logits
$\vz_c^{\mathrm{int}}$, the conditional distribution over the residual logits is given by:
\begin{equation}
\vz_r \mid \vz_c^{\mathrm{int}}
\sim
\mathcal{N}\!\left(
\vmu_r
+
\mSigma_{rc}
\mSigma_{cc}^{-1}
(\vz_c^{\mathrm{int}}-\vmu_c),
\;
\mSigma_{rr}
-
\mSigma_{rc}
\mSigma_{cc}^{-1}
\mSigma_{cr}
\right).
\label{eq:residual_conditioning}
\end{equation}
Consequently, interventions propagate through the learned cross-covariance $\mSigma_{rc}$, updating residual variables statistically dependent on the intervened concepts. The conditioned residual logits are subsequently binarised as in Eq.~\ref{eq:binary_representation} and passed to the task head together with the corrected concepts.

\subsection{Training Objective}
We train Res-SCBM using a base objective $\mathcal{L}_{\mathrm{base}}$ and introduce an additional intervention-aware loss $\mathcal{L}_{\mathrm{int}}$ to encourage the model to remain responsive to concept corrections despite the additional residual capacity.

\paragraph{Base objective.} We adopt the original SCBM training objective \citep{DBLP:conf/nips/VandenhirtzLMV24} and extend it to the joint concept--residual representation. The resulting base objective is given by
\begin{equation}
\begin{aligned}
\mathcal{L}_{\mathrm{base}}
={}&
-\log \frac{1}{M} \sum_{m=1}^{M}
\exp\left(
\sum_{i=1}^{C}
-\operatorname{BCE}\left(\vc_i, \vp_i^{(m)}\right)
\right) \\
&+
\lambda_1
\operatorname{CE}\left(
y,
\frac{1}{M}\sum_{m=1}^{M}
f_\psi\left(
[\hat{\vc}^{(m)},\hat{\vr}^{(m)}]
\right)
\right) \\
&+
\lambda_2
\frac{1}{(C+R)(C+R-1)}
\sum_{i\neq j}
\left|\left[\mSigma(\vx)^{-1}\right]_{ij}\right|.
\end{aligned}
\label{eq:scbm_objective}
\end{equation}
Here, $M$ is the number of Monte Carlo samples and $p_i^{(m)}=\sigma(z_{c,i}^{(m)})$ is the concept probability for sample $m$.
The three terms correspond to the concept prediction loss, downstream task loss and precision loss, respectively. The precision loss regularises the off-diagonal entries of the precision matrix, encouraging a sparse conditional dependency structure and reducing overfitting to spurious dependencies \citep{DBLP:conf/nips/VandenhirtzLMV24}. We retain this regularisation for the joint concept and residual distribution.

\paragraph{Intervention-aware objective.}
While the residual channel can improve predictive performance
under concept incompleteness, it introduces an alternative
pathway from the input to the task prediction. If this pathway
dominates the prediction, concept interventions may have
little effect on the downstream output.

To encourage responsiveness to concept interventions,
we introduce an auxiliary loss $\mathcal{L}_{\mathrm{int}}$
inspired by the intervention objective of
\citet{DBLP:journals/corr/abs-2606-10669}.
This objective penalises task mispredictions when all
observed concepts are intervened on. Specifically, we
replace the predicted concept values with their ground-truth
values $\vc$ and condition the joint Gaussian on the
corresponding corrected concept logits. We then sample
the updated residual representation
$\hat{\vr}^{\mathrm{cond}}_{\vc}$
from the resulting conditional distribution.
The intervention-aware loss is defined as
\begin{equation}
\mathcal{L}_{\mathrm{int}}
=
\mathbb{E}_{(\vx,\vc,y)\sim\mathcal{D}}
\left[
\ell_y\left(
f_\psi([\vc,\hat{\vr}^{\mathrm{cond}}_{\vc}]), y
\right)
\right].
\end{equation}

The combined training objective is
\begin{equation}
\mathcal{L}
=
\mathcal{L}_{\mathrm{base}}
+
\lambda_{\mathrm{int}}\mathcal{L}_{\mathrm{int}},
\label{eq:intResSCBMobj}
\end{equation}
where $\lambda_{\mathrm{int}}$ controls the contribution
of the intervention-aware loss.

By optimising task performance after concept correction, $\mathcal{L}_{\mathrm{int}}$ encourages the model to use corrected concepts together with the conditioned residual representation, thus preserving the influence of concept interventions on the final prediction. It may further promote localisation of concept-specific information within the concept representation \citep{DBLP:journals/corr/abs-2606-10669}, leaving the residual channel to capture complementary task-relevant information beyond the observed concepts.

\section{Experimental setup}
\label{sec:setup}

\paragraph{Datasets.}
We evaluate Res-SCBMs across four vision datasets: \texttt{CUB}, \texttt{CUB-Incomplete}, \texttt{AwA2}, and \texttt{AwA2-Incomplete}. \texttt{CUB}~\citep{WahCUB_200_2011} contains 11,788 images from 200 bird species and provides 312 binary concepts per image. Following \citet{DBLP:conf/icml/KohNTMPKL20}, we retain a subset of 112 concepts. We treat \texttt{CUB} as concept-complete, as the concept set is perfectly predictive of the downstream target.
To evaluate the models under concept incompleteness, we construct \texttt{CUB-Incomplete} following the concept-removal procedure of \citet{DBLP:conf/nips/ZarlengaCDWSJ23}, reducing the concept set from 112 to 22 concepts.
In contrast, \texttt{AwA2} \citep{DBLP:journals/pami/XianLSA19} contains 37,322 images from 50 animal classes, each described by 85 concepts. \texttt{AwA2} has task-specific concept profiles, implying it is a complete concept set.
Hence, we similarly construct \texttt{AwA2-Incomplete} by following the concept-removal procedure of \citet{DBLP:conf/icml/ZarlengaDBSJ25}, reducing the concept set from 85 concepts to an incomplete set with only 9 concepts.
See Appendix~\ref{app:dataset-details} for further details.

\paragraph{Baselines.}
We compare Res-SCBMs against (1) Hard CBMs~\citep{DBLP:conf/nips/HavasiPD22}, an extension of the original CBM with binarised concept predictions, (2) SCBM~\citep{DBLP:conf/nips/VandenhirtzLMV24}, the closest previous work to Res-SCBMs, and (3) CEMs~\citep{DBLP:conf/nips/ZarlengaBCMGDSP22}, a family of leaky models also designed to address incompleteness. Hard CBMs and SCBMs serve as non-leaky baselines, as their binarised concept bottlenecks restrict downstream predictions to the predicted concept values. In contrast, CEM represents each concept with a high-dimensional embedding rather than a single scalar value, providing additional representational capacity. We select CEMs over Soft CBMs as our leaky baseline, given their explicit design to address concept incompleteness while supporting concept interventions. We additionally include a standard DNN as a black-box baseline, providing a reference for predictive performance without the constraints imposed by an interpretable concept bottleneck. Since all models use linear task heads, we also train a linear classifier on ground-truth concepts to assess how much predictive information is available in the concept representation alone. For comparability with our intervention-aware model, and following \citet{DBLP:conf/nips/ZarlengaBCMGDSP22}, we train CEMs by randomly intervening on $25\%$ of the concepts during training (a process the authors call \textit{RandInt}). We retain \textit{RandInt} rather than explicitly optimising $\mathcal{L}_{\mathrm{int}}$, as \citet{DBLP:journals/corr/abs-2606-10669} show that both approaches can yield intervenable CEMs, with \textit{RandInt} implicitly optimising a similar objective.\looseness-1

\paragraph{Metrics.}
We evaluate predictive performance using task accuracy and concept prediction performance using mean concept ROC-AUC. To evaluate intervenability, we measure task accuracy w.r.t.\ the fraction of corrected concepts (selected at random). This yields intervention curves that quantify how effectively improvements to the concept representation propagate to the downstream prediction.

\paragraph{Implementation Details.}
For all residual models, we use a residual dimension of 20 unless stated otherwise. For the training objectives of our models, as defined in Equations~\ref{eq:scbm_objective} and~\ref{eq:intResSCBMobj}, we set $\lambda_1=\lambda_2=\lambda_{\mathrm{int}}=1$. We use amortised covariance parametrisation for SCBMs and Res-SCBMs and a random intervention policy for all models. Further implementation details are provided in Appendix~\ref{app:architecture}.

\section{Results}
\label{sec:results}
We evaluate two variants: Res-SCBM w/o $\mathcal{L}_{\mathrm{int}}$, trained using only the base objective, and Res-SCBM, which
additionally incorporates the intervention-aware loss $\mathcal{L}_{\mathrm{int}}$.
We address four main research questions:
\begin{enumerate}[label=\textbf{RQ\arabic*.}, leftmargin=*, topsep=0pt, itemsep=-1pt]
    \item Do our residual models improve predictive performance compared with models restricted to the concept representation?
    \item Do our residual models retain effective concept interventions despite the added capacity?
    \item Does intervention-aware training preserve intervenability as residual capacity increases?
    \item What do the learned concept–residual covariances reveal about the residual representation?
\end{enumerate}

\subsection{Predictive Performance (RQ1)}
\label{sec:performance-experiment}

We first look at Res-SCBM's task accuracy across tasks when no interventions are performed. Our results summarised in Table~\ref{tab:target-results}, show that, on concept-complete tasks, Res-SCBM is slightly stronger than concept-based baselines, while on concept-incomplete tasks, Res-SCBM substantially outperforms non-leaky baselines (i.e., Hard CBMs and SCBMs). This latter improvement is achieved while maintaining a competitive, and sometimes better (e.g., on \texttt{CUB-Incomplete}), accuracy than high-capacity CEMs.
We see this even when we train Res-SCBMs without $\mathcal{L}_\text{int}$: for example, on \texttt{AwA2-Incomplete}, both variants of Res-SCBM improve task accuracy by more than 23 percentage points over Hard CBM and SCBM, while performing comparably to CEMs. Crucially, both Res-SCBM variants outperform a linear classifier trained to map ground-truth observed concepts to the task labels on \texttt{AwA2-Incomplete}, suggesting that their predictions exploit task-relevant information beyond the incomplete concept set.
Finally, we observe that Res-SCBMs are also competitive w.r.t.\ unconstrained DNNs: Res-SCBM outperforms the black-box DNN on both \texttt{CUB} tasks and performs similarly to both \texttt{AwA2} tasks. On the concept-complete datasets, the gains over existing CBMs are comparatively small, consistent with the observed concepts already containing sufficient information for the downstream task.\looseness-1

\begin{table}[!t]
    \centering
    \caption{
        Task accuracy (\%) before concept interventions. Values are means over seeds, with standard deviations in subscript. Best/second-best concept-based results are bold/underlined. The DNN is reported as a predictive baseline, while the linear head uses ground-truth concepts and is reported as a reference. Residual models outperform non-leaky baselines under concept
incompleteness.
    }
    \label{tab:target-results}
    
    \setlength{\tabcolsep}{4pt}
    \begin{tabular}{lcccc}
        \toprule
        Method
        & \texttt{CUB}
        & \shortstack{CUB\\Incomplete}
        & \texttt{AwA2}
        & \shortstack{AwA2\\Incomplete} \\
        \midrule

        DNN (black box)
        & $68.60_{\pm 0.44}$
        & $64.77_{\pm 0.15}$
        & $92.93_{\pm 0.15}$
        & $92.50_{\pm 0.10}$ \\

        Linear head (GT concepts)
        & $100.00_{\hspace{9mm}}$
        & $84.91_{\pm 0.72}$
        & $100.00_{\hspace{9mm}}$
        & $76.20_{\pm 0.03}$ \\

        \midrule

        Hard CBM
        & $69.63_{\pm 0.90}$
        & $47.87_{\pm 1.00}$
        & $92.30_{\pm 0.10}$
        & $68.20_{\pm 0.87}$ \\

        CEM
        & $70.50_{\pm 0.36}$
        & \underline{$64.77_{\pm 1.71}$}
        & \underline{$92.63_{\pm 0.25}$}
        & $\mathbf{91.87}_{\boldsymbol{\pm 0.06}}$ \\

        SCBM
        & \underline{$71.30_{\pm 0.00}$}
        & $49.03_{\pm 0.71}$
        & $91.87_{\pm 0.15}$
        & $68.43_{\pm 0.31}$ \\

        \midrule

        Res-SCBM w/o $\mathcal{L}_{\mathrm{int}}$ (ours)
        & $70.87_{\pm 0.15}$
        & $64.43_{\pm 0.72}$
        & $92.60_{\pm 0.17}$
        & $\mathbf{91.87}_{\boldsymbol{\pm 0.06}}$ \\

        Res-SCBM (ours)
        & $\mathbf{71.80}_{\boldsymbol{\pm 0.70}}$
        & $\mathbf{68.20}_{\boldsymbol{\pm 0.69}}$
        & $\mathbf{92.77}_{\boldsymbol{\pm 0.06}}$
        & \underline{$91.77_{\pm 0.12}$} \\

        \bottomrule
    \end{tabular}
\end{table}

Similar trends can be observed in concept prediction performance. Specifically, as shown in Table~\ref{tab:concept-auroc},  the mean concept ROC-AUC of Res-SCBMs remains competitive compared with previous baselines. This holds for complete and incomplete tasks, suggesting that the residual pathway in Res-SCBMs may indeed capture complementary task-relevant information not in the concept set.

\begin{table}[!t]
    \centering
    \caption{
        Mean concept ROC-AUC (\%) before concept interventions.
        Values are means over seeds, with standard deviations in subscript.
        Best/second-best concept-based results are bold/underlined. Residual models remain competitive with baseline concept prediction performance.
    }
    \label{tab:concept-auroc}

    \setlength{\tabcolsep}{4pt}
    \begin{tabular}{lcccc}
        \toprule
        Method
        & \texttt{CUB}
        & \shortstack{CUB\\Incomplete}
        & \texttt{AwA2}
        & \shortstack{AwA2\\Incomplete} \\
        \midrule

        Hard CBM
        & \underline{$97.83_{\pm 0.12}$}
        & $97.00_{\pm 0.17}$
        & \underline{$99.30_{\pm 0.00}$}
        & \underline{$99.10_{\pm 0.00}$} \\

        CEM
        & \underline{$97.83_{\pm 0.06}$}
        & $97.13_{\pm 0.15}$
        & $\mathbf{99.37}_{\boldsymbol{\pm 0.06}}$
        & $\mathbf{99.27}_{\boldsymbol{\pm 0.06}}$ \\

        SCBM
        & $\mathbf{97.97}_{\boldsymbol{\pm 0.06}}$
        & $97.17_{\pm 0.06}$
        & $99.17_{\pm 0.06}$
        & \underline{$99.10_{\pm 0.00}$} \\

        \midrule

        Res-SCBM w/o $\mathcal{L}_{\mathrm{int}}$ (ours)
        & $\mathbf{97.97}_{\boldsymbol{\pm 0.06}}$
        & \underline{$97.33_{\pm 0.06}$}
        & $99.00_{\pm 0.00}$
        & \underline{$99.10_{\pm 0.00}$} \\

        Res-SCBM (ours)
        & $\mathbf{97.97}_{\boldsymbol{\pm 0.06}}$
        & $\mathbf{97.53}_{\boldsymbol{\pm 0.12}}$
        & $99.03_{\pm 0.06}$
        & \underline{$99.10_{\pm 0.00}$} \\

        \bottomrule
    \end{tabular}
\end{table}

\subsection{Test-Time Concept Interventions (RQ2)}
\label{sec:intervention-experiments}

\begin{figure}[!t]
    \centering
    \includegraphics[width=\linewidth]{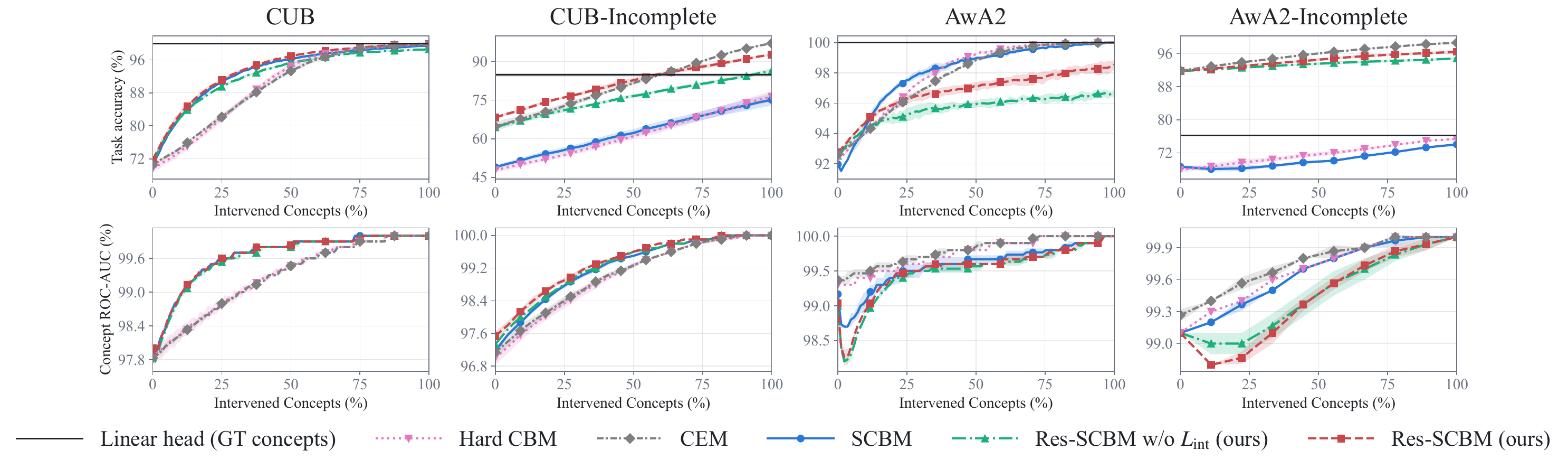}
    \vspace*{-5mm}
    \caption{Task accuracy (top) and mean concept ROC-AUC (bottom) as the fraction of corrected concepts increases on \texttt{CUB}, \texttt{CUB-Incomplete}, \texttt{AwA2}, \texttt{AwA2-Incomplete}. Res-SCBM remains responsive to concept corrections despite its residual pathway.}
    \label{fig:intervention_curves}
\end{figure}

Next, we examine how interventions affect the downstream task accuracies and concept ROC-AUC of Res-SCBMs.  Our results, summarised in Figure~\ref{fig:intervention_curves}, show that, across all datasets, Res-SCBM remains responsive to concept corrections despite its additional residual pathway.
For example, in \texttt{CUB-Incomplete} we see that Res-SCBM's task accuracy improves substantially as more concepts are corrected, reaching above 90\% task accuracy under full intervention and approaching CEM's intervention performance. Notably, Res-SCBM outperforms CEM when only a few concepts are corrected. This is particularly beneficial in practice, where the cost and effort of obtaining ground-truth concept corrections often limit the number of feasible interventions.

Crucially, propagating interventions across Res-SCBM's bottleneck increases its concept ROC-AUC above that of CEMs on both \texttt{CUB} datasets, with a marked improvement on the complete version. On both \texttt{AwA2} datasets, Res-SCBM shows a slight initial decrease in concept ROC-AUC before recovering as more concepts are corrected. A similar trend is observed for SCBM on complete \texttt{AwA2}, indicating that this behaviour is not unique to the residual extension. On complete \texttt{AwA2}, Res-SCBM also remains below the ground-truth-concept reference under full intervention. Nevertheless, Res-SCBM is designed primarily for concept-incomplete settings, where it achieves strong predictive performance while retaining responsiveness to concept interventions.

Finally, we observe that Res-SCBM is also consistently more responsive than Res-SCBM w/o $\mathcal{L}_{\mathrm{int}}$, with the largest separation on \texttt{CUB-Incomplete} and a smaller but consistent difference on \texttt{AwA2-Incomplete}. This suggests that intervention-aware training increases the influence of corrected concept information on the downstream prediction.
Together, these results suggest that, at the default residual capacity, both variants remain responsive to concept corrections, while intervention-aware training strengthens this response.

\subsection{Effect of Intervention-Aware Training (RQ3)}
\label{sec:int-aware-experiment}

\begin{figure}[!t]
    \centering
    \includegraphics[width=\linewidth]{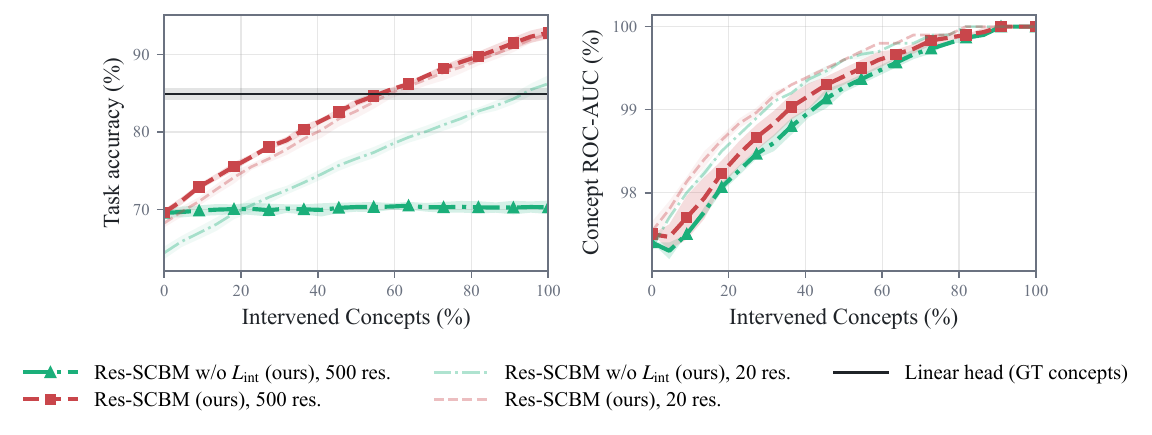}
    \vspace*{-6mm}
    \caption{\textbf{Effect of intervention-aware training.}
    Task accuracy (left) and mean concept ROC-AUC (right) under concept interventions on \texttt{CUB-Incomplete}, with/without  $\mathcal{L}_{\mathrm{int}}$ and residual dimensions $R \in \{20, 500\}$. The horizontal line shows linear classifier accuracy on ground-truth concepts. Intervention-aware training preserves responsiveness at increased residual capacity.}
    \label{fig:intervention-curves-500-res}
\end{figure}

To examine how intervention-aware training interacts with residual capacity, we compare \mbox{Res-SCBM} with and without $\mathcal{L}_{\mathrm{int}}$ using residual dimensions $R=20$ and $R=500$. The larger residual dimension serves as a stress test in which the residual pathway has substantially greater capacity to provide an alternative route to the downstream prediction.
Our results, shown in Figure~\ref{fig:intervention-curves-500-res}, suggest that residual capacity has a pronounced effect when we remove intervention-aware training. Specifically, we observe that, while Res-SCBM w/o $\mathcal{L}_{\mathrm{int}}$ remains responsive to concept corrections when $R=20$, increasing the residual dimension to $R=500$ leaves task accuracy almost unchanged as additional concepts are corrected. In contrast, Res-SCBM remains strongly responsive for both $R=20$ and $R=500$, with similar intervention trajectories across both.

Importantly, the concept ROC-AUC trajectories are similar across the model variants. This suggests that the divergence in task accuracy reflects how effectively corrected concept information influences the downstream prediction, rather than differences in concept prediction quality. Moreover, Res-SCBM eventually exceeds the linear classifier using ground-truth observed concepts, suggesting that the model combines corrected concept information with complementary information retained in the residual representation.

These results indicate that increasing residual capacity can substantially weaken intervenability when the model is trained without an intervention-aware objective. The effect of $\mathcal{L}_{\mathrm{int}}$ therefore becomes particularly important as residual capacity increases, preserving the influence of concept corrections despite a high-capacity residual pathway.

\subsection{Concept--Residual Covariance Analysis (RQ4)}
\label{sec:residual-experiments}

\begin{figure}[!t]
    \centering
    \includegraphics[width=0.7\linewidth]{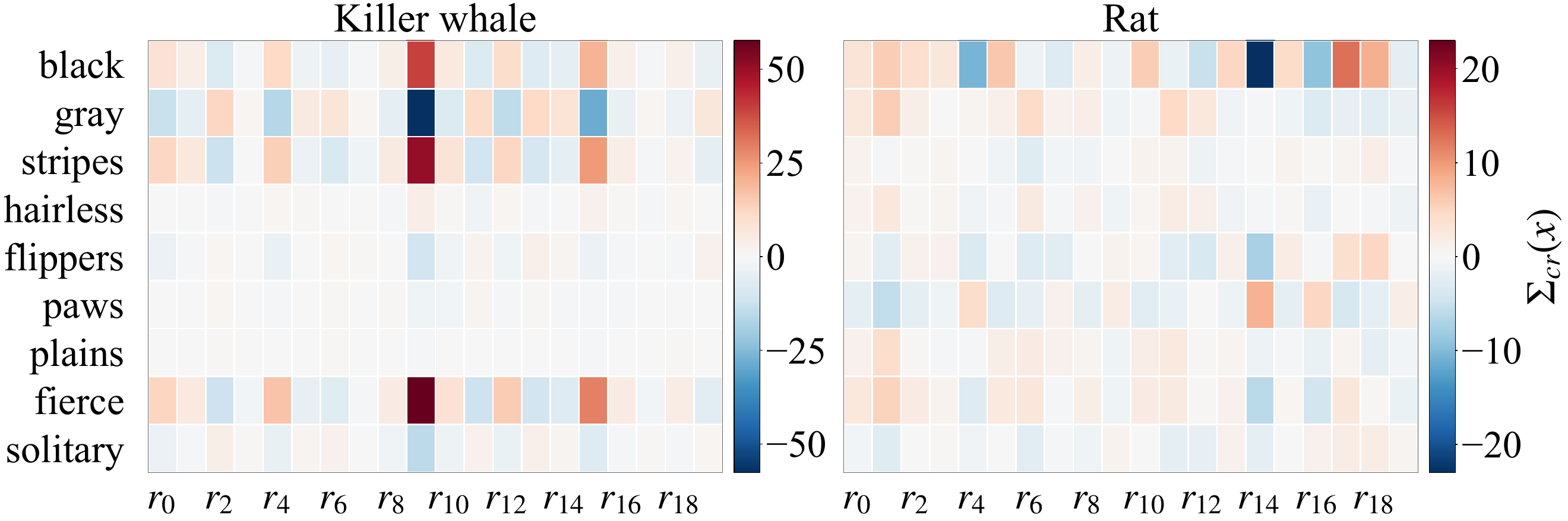}
    \caption{Learned concept–residual cross-covariance for two \texttt{AwA2-Incomplete} test samples (concepts: rows; residuals: columns). Distinct patterns across residual dimensions and samples reveal input-dependent concept–residual dependencies.}
    \label{fig:cross-covariance}
\end{figure}

Figure \ref{fig:cross-covariance} illustrates the learned concept–residual cross-covariance for two \texttt{AwA2-Incomplete} test samples. Different residual dimensions exhibit distinct dependency patterns with the observed concepts, which also vary across inputs. These dependencies are central to Res-SCBM's intervention mechanism, as Gaussian conditioning uses the learned covariance structure to update statistically dependent residual variables following concept corrections. Inspecting the cross-covariance therefore provides a mechanistic perspective on how concepts and residuals interact and which residual dimensions may respond to particular concept corrections. Comparing these dependencies across samples could also help infer what individual residual dimensions might encode. A more detailed investigation of these relationships and their potential for improving residual interpretability is left for future work.

\section{Limitations and Conclusion}

\paragraph{Limitations \& Future Work.}
As in SCBMs, modelling a joint Gaussian distribution requires learning a covariance matrix whose size scales quadratically with the number of variables. Moreover, the effectiveness of intervention propagation depends on the learned concept--residual cross-covariance. Concept and residual predictions may covary across inputs through their predicted means without substantial within-input cross-covariance, limiting the effect of Gaussian conditioning on the residual representation. Future work could therefore investigate methods to encourage stronger concept--residual cross-covariance to improve intervention propagation. A natural extension would also be to leverage the learned covariance structure to better understand what the residual channel encodes and how it interacts with observed concepts.



\paragraph{Conclusion.}
We introduced Res-SCBMs, a concept-based architecture for addressing concept incompleteness while preserving effective test-time interventions. Res-SCBMs augment the observed concept representation with residual variables and jointly model both through a multivariate Gaussian distribution. The learned concept--residual covariance allows concept interventions to propagate to statistically dependent residual variables through Gaussian conditioning. This joint representation also offers  future research directions, including exploiting the learned dependencies to better understand the information captured by residual variables and potentially discover missing concepts.
Our experiments show that Res-SCBMs improve predictive performance on concept-incomplete tasks while maintaining effective test-time interventions. Importantly, our model's intervention-aware training helps preserve responsiveness to concept corrections even when the residual representation has substantially greater capacity. These findings demonstrate that additional residual capacity can be combined with effective concept interventions, providing a step towards concept-based models that remain both predictive and intervenable in real-world incomplete settings.

\newpage

\subsection*{AI use statement}
In this work, we used generative AI tools to review the manuscript for clarity, grammar, and readability. Generative AI was also used to identify relevant literature, with every identified paper manually read and evaluated by the authors. Generative AI was used to assist with code development, primarily by generating skeleton code, assisting with plot formatting, and identifying areas of the codebase that could be cleaned up or reorganised. Additionally, generative AI was consulted to check aspects of the proposed methodology. These checks were independently verified by the authors through a manual review of the relevant literature. All AI-assisted code was manually inspected and tested for correctness, and all plots were manually inspected and verified. The manuscript was written, thoroughly reviewed, and revised by the authors.

Generative AI was not used to develop theoretical models or conceptual frameworks, formulate mathematical claims, provide critical ingredients for mathematical proofs, or assist in writing proofs. It was not used to propose or refine research hypotheses, design experiments, or develop the research methodology beyond the methodological checks described above. No generative AI tools were used for translation, dataset cleaning or reformatting, qualitative or thematic data analysis, or interpretation of experimental results.

All research ideas, experimental design, execution of experiments, analysis, and interpretation of results were carried out by the authors. All AI-assisted material was reviewed, and the authors take full responsibility for the final content of this work, including its results, claims, and conclusions.

\subsection*{Reproducibility statement}

To support reproducibility, we provide a detailed description of our model formulation, intervention procedure, and training objectives in Section~\ref{sec:model-formulation}. Appendix~\ref{app:dataset-details} describes the datasets, preprocessing, and construction of the concept-incomplete variants. Appendix~\ref{app:implementation} provides implementation details and computational resources, while Appendix~\ref{app:architecture} documents the model architectures, training procedures, hyperparameters, and test-time intervention protocols. All experiments were conducted using three random seeds, with results reported as mean and standard deviation. Our implementation is publicly available on GitHub.\footnote{\url{https://github.com/stephenmarcou/residual_scbm}}



\subsubsection*{Acknowledgments}
 
MV and SL are supported by the Swiss State Secretariat for Education, Research, and Innovation (SERI) under contract number MB22.00047.

\bibliography{references_check}
\bibliographystyle{iclr2027_conference}

\newpage
\appendix
\section{Appendix}

\subsection{Dataset Details}
\label{app:dataset-details}
\paragraph{CUB.}
The Caltech-UCSD Birds-200-2011 (CUB)~\citep{WahCUB_200_2011} dataset comprises 11,788 images across 200 bird classes, annotated with 312 binary attributes organised into 28 concept groups. Following the preprocessing procedure of~\citet{DBLP:conf/icml/KohNTMPKL20}, we retain 112 binary class-level concepts, where all images belonging to the same bird class share the same ground-truth concept labels.

\paragraph{CUB-Incomplete.}
Following~\citet{DBLP:conf/nips/ZarlengaCDWSJ23}, we construct
\texttt{CUB-Incomplete} by retaining 25\% of the original concept groups from \texttt{CUB}.
This results in 22 binary concepts across seven groups:
bill shape, head pattern, breast colour, bill length,
wing shape, tail pattern, and bill colour.
The dataset retains all 200 bird classes and uses the
same train, validation, and test splits as \texttt{CUB}. 

\paragraph{AwA2.}
We use the Animals with Attributes 2 (AwA2) dataset
\citep{DBLP:journals/pami/XianLSA19}, comprising
37,322 images across 50 animal classes. Each class is annotated with 85 binary attributes. Following~\citet{DBLP:conf/icml/ZarlengaDBSJ25}, the attributes are organised into 28 concept groups. Unlike instance-level annotations, the attributes are defined
at the class level, meaning that all images belonging to the
same animal class share the same ground-truth concept labels.

\paragraph{AwA2-Incomplete.}
Following~\citet{DBLP:conf/icml/ZarlengaDBSJ25},
we use a concept-incomplete variant of \texttt{AwA2} that retains
nine of the original 85 concepts:
black, gray, stripes, hairless, flippers, paws, plains,
fierce, and solitary. The dataset retains all 50 animal
classes and uses the same train, validation, and test
splits as \texttt{AwA2}.

\paragraph{Dataset overview.}
Table~\ref{tab:dataset-properties} summarises the
properties of all datasets used in our experiments.

\begin{table}[h]
    \centering
    \caption{High-level properties of the datasets used in our
    experiments. Train, val., and test indicate the number of
    samples in each split. All input images have dimensions
    $3 \times 224 \times 224$.}
    \label{tab:dataset-properties}
    \begin{tabular}{@{}lrrrrrr@{}}
        \toprule
        Dataset & Train & Val. & Test
        & Classes & Concepts & Groups \\
        \midrule
        \texttt{CUB}
        & 4,796 & 1,198 & 5,794 & 200 & 112 & 28 \\
        \texttt{CUB-Incomplete}
        & 4,796 & 1,198 & 5,794 & 200 & 22 & 7 \\
        \texttt{AwA2}
        & 22,393 & 7,464 & 7,465 & 50 & 85 & 28 \\
        \texttt{AwA2-Incomplete}
        & 22,393 & 7,464 & 7,465 & 50 & 9 & 6 \\
        \bottomrule
    \end{tabular}
\end{table}

\subsection{Implementation Details}
\label{app:implementation}
All models were implemented in PyTorch v2.4.1 and trained on a GPU cluster equipped with NVIDIA GeForce RTX 4090 GPUs. Each training job was allocated one GPU and two CPU cores. All experiments were conducted using three random seeds, and results are reported as mean $\pm$ standard deviation across seeds.

\subsection{Model Architecture and Hyperparameters}
\label{app:architecture}

\paragraph{Architecture.}
For the \texttt{CUB}-based datasets, we use an ImageNet-pretrained ResNet-18 \citep{DBLP:conf/cvpr/HeZRS16} as the image backbone. For the \texttt{AwA2}-based datasets, we use frozen, 2048-dimensional embeddings extracted using an ImageNet-pretrained ResNet-101, following \citet{DBLP:journals/pami/XianLSA19}. All models use a linear downstream task head.

\paragraph{SCBM and Res-SCBM.}
The original SCBM formulation considers both global and amortised covariance parametrisations \citep{DBLP:conf/nips/VandenhirtzLMV24}. The global variant learns a single covariance matrix shared across the dataset, whereas the amortised variant predicts an input-dependent covariance matrix for each sample. For SCBM and both Res-SCBM variants, we use linear output heads to parameterise the Gaussian distribution and employ the amortised covariance parametrisation throughout. We use $M=100$ Monte Carlo samples to approximate the training objectives. The loss coefficients defined in Eqs~\ref{eq:scbm_objective} and~\ref{eq:intResSCBMobj} are set to $\lambda_1=\lambda_2=\lambda_{\mathrm{int}}=1$, where applicable.

Unless otherwise specified, we use $R=20$ residual variables across all experiments. For the experiment investigating the effect of intervention-aware training in Section~\ref{sec:int-aware-experiment} under increased residual capacity, we increase the residual dimension to $R=500$. Both Res-SCBM variants share the same architecture and hyperparameters, differing only in whether the intervention-aware loss $\mathcal{L}_{\mathrm{int}}$ is included.\looseness-1

\paragraph{CEM.}
For CEM, we use an embedding dimension of $16$ per concept, following \citet{DBLP:conf/nips/ZarlengaBCMGDSP22}. We additionally employ random intervention training (RandInt) with an intervention probability of $0.25$, where randomly selected concepts are corrected to their ground-truth values during training~\citep{DBLP:conf/nips/ZarlengaBCMGDSP22}.

\paragraph{DNN (black box).}
We use a black-box classifier with the same encoder as the corresponding concept-based models, followed by a linear layer, a LeakyReLU activation, and a linear classification head. To match the bottleneck capacity of both Res-SCBM variants, we set the hidden dimension to $d=C+20$, where $C$ is the number of observed concepts and 20 is the residual dimension. Unlike the concept-based models, all hidden units are unsupervised, and the DNN is trained solely using task cross-entropy.

\paragraph{Training.}
All models were trained for a maximum of 200 epochs, with early stopping patience of 75 epochs for \texttt{CUB} and \texttt{CUB-Incomplete} and 30 epochs for \texttt{AwA2} and \texttt{AwA2-Incomplete}. We use a learning rate of $10^{-4}$ and a batch size of 64 across all models.

\paragraph{Concept interventions.}
For all test-time intervention experiments, we use a random intervention policy~\citep{DBLP:conf/icml/KohNTMPKL20,DBLP:conf/nips/VandenhirtzLMV24}, selecting a random subset of concepts for correction at each intervention step.

For SCBM and both Res-SCBM variants, we compute corrected concept logits using the empirical percentile strategy described in Eq.~\ref{eq:intervention_logits}. We then condition the joint Gaussian distribution on these logits, allowing interventions to propagate to statistically dependent concepts and, in our models, residual variables.

For Hard CBM~\citep{DBLP:conf/nips/HavasiPD22} and CEM~\citep{DBLP:conf/nips/ZarlengaBCMGDSP22}, we instead replace the predicted concept probabilities with their ground-truth binary values. In Hard CBM, this corrects the corresponding entries in the binary concept bottleneck. In CEM, it selects the corresponding negative or positive concept embedding instead of the original probability-weighted combination of the two.\looseness-1

\end{document}